\documentclass[conference]{IEEEtran}
\IEEEoverridecommandlockouts
\usepackage{cite}
\usepackage{amsmath,amssymb,amsfonts}
\usepackage{algorithmic}
\usepackage{graphicx}
\usepackage{textcomp}
\usepackage{xcolor}
\usepackage{hyperref}
\usepackage{float}
\usepackage{xcolor}
\usepackage{tikz}
\usepackage{subcaption}
\usepackage{pgf-pie}   
\usepackage{pgfplots}
\usepackage{pgfplotstable}
\usepackage[table]{xcolor}
\usepackage{arydshln}
\usepackage{multirow}
\usepackage{booktabs}
\PassOptionsToPackage{table}{xcolor}
\usepackage{tcolorbox}
\usetikzlibrary{calc}
\pgfplotsset{compat=1.18}
\usepgfplotslibrary{ternary}
\usepackage[export]{adjustbox}
\usepgfplotslibrary{polar}
\usepgfplotslibrary{groupplots}

\def\BibTeX{{\rm B\kern-.05em{\sc i\kern-.025em b}\kern-.08em
    T\kern-.1667em\lower.7ex\hbox{E}\kern-.125emX}}
\begin{document}

\title{RLAIF-SPA: Structured AI Feedback for Semantic-Prosodic Alignment in Speech Synthesis}


\author{
Qing Yang$^{1}$, Zhenghao Liu$^{1}$\textsuperscript{\textdagger}, Yangfan Du$^{1}$, Pengcheng Huang$^{1}$, Tong Xiao$^{1}$\textsuperscript{\textdagger}\\
$^{1}$School of Computer Science and Engineering, Northeastern University, China
}

\maketitle

\begingroup
\renewcommand{\thefootnote}{\fnsymbol{footnote}}
\footnotetext[2]{Corresponding author.}
\endgroup
\setcounter{footnote}{0}

\begin{abstract}
Recent advances in Text-To-Speech (TTS) synthesis have achieved near-human speech quality in neutral speaking styles. However, most existing approaches either depend on costly emotion annotations or optimize surrogate objectives that fail to adequately capture perceptual emotional quality. As a result, the generated speech, while semantically accurate, often lacks expressive and emotionally rich characteristics.
To address these limitations, we propose RLAIF-SPA, a novel framework that integrates Reinforcement Learning from AI Feedback (RLAIF) to directly optimize both emotional expressiveness and intelligibility without human supervision. Specifically, RLAIF-SPA incorporates Automatic Speech Recognition (ASR) to provide semantic accuracy feedback, while leveraging structured reward modeling to evaluate prosodic–emotional consistency.
RLAIF-SPA enables more precise and nuanced control over expressive speech generation along four structured evaluation dimensions: Structure, Emotion, Speed, and Tone. Extensive experiments on LibriSpeech, MELD, and Mandarin ESD datasets demonstrate consistent gains across clean read speech, conversational dialogue, and emotional speech. On LibriSpeech, RLAIF-SPA consistently outperforms Chat-TTS, achieving a 26.1\% reduction in word error rate, a 9.1\% improvement in SIM-O, and over 10\% gains in human subjective evaluations.
\end{abstract}

\begin{IEEEkeywords}
Emotional Speech Synthesis, Reinforcement Learning, AI Feedback
\end{IEEEkeywords}

\section{Introduction}
Recent advances in Large Language Models (LLMs) have enabled Text-To-Speech (TTS) systems to achieve near-human naturalness and intelligibility \cite{barakat2024deep,liu2026waveex, chen2023vector}. Nevertheless, human interaction is not solely about informational exchange but also involves the subtle expression of emotion. These emotional signals play a crucial role in shaping listener engagement and comprehension, particularly in applications such as conversational agents, audiobooks, and virtual assistants. In these settings, insufficient emotional expressiveness often leads to speech that sounds monotonous or flat, thereby reducing its effectiveness in accurately conveying human emotions \cite{park2024avemodialogchataudiovisualusers}. Thus, developing TTS models with controllable emotional capabilities is essential for bridging this gap and has become a major focus in the field.

Despite this progress, emotional TTS still faces a core bottleneck: scalable supervision for controllable expressiveness while preserving content fidelity. Label-conditioned methods offer an intuitive control interface, but coarse emotion categories fail to capture fine-grained prosody and scaling typically relies on costly human annotations. Preference-based reinforcement learning provides another direction by optimizing perceptual quality, yet a single holistic score offers weak credit assignment and does not reveal which prosodic attributes are misaligned, making targeted control difficult. Meanwhile, improving expressiveness often degrades intelligibility when content fidelity is not explicitly constrained. These challenges call for a framework that provides structured, attribute-level feedback at scale and jointly optimizes expressiveness and content fidelity.

\begin{figure*}[t]
    \centering
    \includegraphics[width=1\textwidth,keepaspectratio=true, trim=0cm 0cm 0cm 0cm,clip]{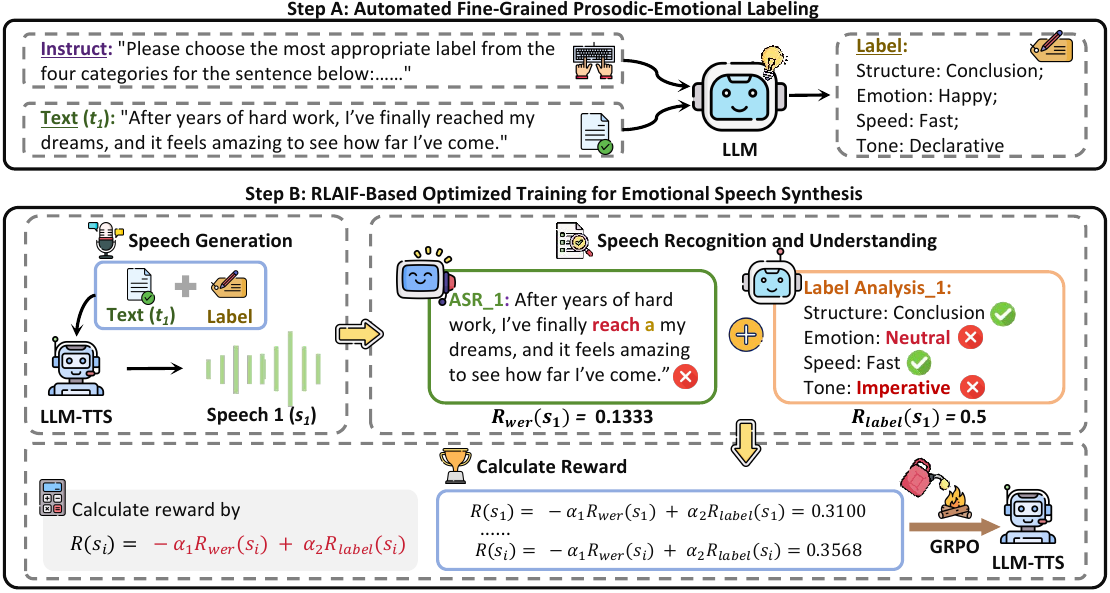} 
    \caption{Illustration of the proposed RLAIF-SPA framework. RLAIF-SPA optimizes both emotional expressiveness and accuracy through the AI Feedback mechanism, which is generated by AI models themselves.}
    \label{fig:overview}
    \vspace{-12pt}
\end{figure*}

We view emotional TTS post-training as a multi-attribute alignment problem where holistic preference scores provide weak credit assignment, motivating diagnostic AI feedback at the attribute level. To address these challenges, we propose \textbf{RLAIF-SPA}, a framework that incorporates \textbf{R}einforcement \textbf{L}earning from \textbf{AI} \textbf{F}eedback (RLAIF) \cite{bai2022constitutional} for \textbf{S}emantic-\textbf{P}rosodic \textbf{A}lignment in Text-To-Speech synthesis. Unlike prior RLHF-style post-training that optimizes a holistic preference, RLAIF-SPA introduces a structured AI-feedback interface that decomposes alignment into semantic fidelity and attribute-wise prosodic alignment, enabling joint optimization of expressiveness and accuracy with automatically computed rewards. RLAIF-SPA leverages two core feedback components. For expressiveness, \textit{Prosodic Label Alignment} judges the model output against automatically generated labels along four fine-grained dimensions: \textit{Structure, Emotion, Speed,} and \textit{Tone} \cite{ji2025controlspeechsimultaneousindependentzeroshot}. For clarity, \textit{Semantic Accuracy Feedback} assesses consistency between the transcribed output and the original input text. By combining these two signals, the AI Feedback mechanism provides a stable and scalable optimization target. 

Experiments on LibriSpeech, MELD, and ESD show that RLAIF-SPA achieves a better trade-off between accuracy and expressiveness than strong baselines, reducing word error rate while improving speaker similarity and emotion alignment. Beyond improved metrics, our key contribution is a structured AI-feedback interface that decomposes emotional TTS post-training into attribute-wise prosodic alignment and semantic accuracy supervision, enabling diagnostic credit assignment and preventing expressiveness gains from degrading content fidelity. Our code is released in an repository: \url{https://github.com/Zoe-Mango/RLAIF-SPA}.

\section{Related Work}
Early emotional TTS systems typically condition synthesis on explicit, coarse-grained emotion labels \cite{diatlova2023emospeechguidingfastspeech2emotional,guo2023emodiffintensitycontrollableemotional,kang2023zetspeechzeroshotadaptiveemotioncontrollable}. By associating discrete categories with acoustic and prosodic patterns, these methods modulate speech attributes to elicit the desired affect \cite{tang2023emomixemotionmixingdiffusion}. Despite their simplicity and controllability, this paradigm faces two inherent limitations. First, discrete labels oversimplify the continuous and compositional nature of human affect, making it difficult to represent subtle intensity changes or mixed emotions. Second, learning high-quality emotional control at scale typically relies on extensive manual annotations, rendering large-scale corpus construction labor-intensive and costly and thus limiting scalability \cite{um2019emotionalspeechsynthesisrich,guan2024mm,wu2024laugh}.

To reduce dependence on categorical emotion labels and manual annotation, recent studies have explored Reinforcement Learning (RL) paradigms that optimize perceptual objectives from human or model-based feedback \cite{rafailov2024directpreferenceoptimizationlanguage,lin2025align, LiaoHYD0XXLL024}. Leveraging pairwise preference annotations, these approaches can model human judgments of emotional expressiveness and produce more nuanced and diverse speech \cite{gao2025emo,zhang2024speechalignaligningspeechgeneration,cideron2024musicrlaligningmusicgeneration}. However, existing RLHF-based TTS methods predominantly rely on holistic preference signals. In practice, optimizing a single scalar score entangles multiple factors and provides limited diagnostic guidance on which prosodic dimensions should be adjusted\cite{ge2026emotion}. As a result, these models often exhibit limited controllability and struggle to support precise, targeted optimization for emotional speech synthesis, especially when expressiveness improvements risk degrading semantic fidelity.

Expressive speech is inherently multi-dimensional and is therefore better modeled through complementary prosodic attributes\cite{larrouy2025sound}. Speaking rate and pitch variation correlate strongly with arousal and subtle affective cues \cite{KALLINEN2004275,Cho2024}, while higher-level structures such as discourse rhythm contribute to contextual coherence \cite{Galdino2025,ma2024emoboxmultilingualmulticorpusspeech}. These findings motivate decomposing emotional alignment into attribute-wise supervision rather than relying on a single overall preference. In parallel, group-relative formulations such as Group Relative Policy Optimization (GRPO) compare multiple candidates for the same input \cite{shao2024deepseekmath, liu2025group}, yielding a more stable learning signal than noisy absolute scoring. RLAIF-SPA builds on these insights by introducing structured, attribute-level AI feedback and jointly constraining semantic fidelity, enabling scalable and controllable emotional TTS without requiring human preference labels.

\section{Methodology}
This section presents the proposed RLAIF-SPA framework. We first describe the overall training strategy and optimization objectives, and then introduce the two core components of the AI feedback mechanism, Prosodic Label Alignment and Semantic Accuracy Feedback, as illustrated in Fig.~\ref{fig:overview}.

\subsection{Training Strategy and Optimization Framework}
The training of RLAIF-SPA is formulated as a multi-objective optimization problem guided by feedback provided by an automated evaluator. Specifically, the feedback signals correspond to two key objectives: emotional expressiveness and semantic accuracy. The optimization aims to improve prosodic-emotional label alignment while simultaneously reducing the Word Error Rate (WER), thereby encouraging the model to generate speech that is both emotionally expressive and accurate.

Given an input text $t_i$, we first construct a target prosodic label vector $\mathbf{y}_i$ with four dimensions, automatically generated by a LLM. The policy model $\pi_\theta$ then produces a speech sample $s_i \sim \pi_\theta(\cdot \mid t_i)$, which is scored by a composite reward function $R(s_i)$ capturing two complementary objectives: (i) prosodic-emotional alignment to $\mathbf{y}_i$, evaluated by an external speech understanding model, and (ii) semantic accuracy with respect to $t_i$, quantified using a transcription produced by an Automatic Speech Recognition (ASR) system. These signals are combined through a weighted linear form:

\begin{equation}
R(s_i) = - \alpha_1 R_{\text{wer}}(s_i) + \alpha_2 R_{\text{label}}(s_i),
\label{eq:total_reward}
\end{equation}
where $R_{\text{label}}(s_i)$ denotes the reward for prosodic-emotional alignment (Section~\ref{sec:emotion_alignment}), and $R_{\text{wer}}(s_i)$ is a penalty term based on the WER of the generated speech (Section~\ref{sec:text_alignment}). The non-negative hyperparameters $\alpha_1$ and $\alpha_2$ control the trade-off between accuracy and emotional expressiveness.

To optimize the policy model, we adopt Group Relative Policy Optimization (GRPO)~\cite{shao2024deepseekmath}, which evaluates the relative quality of multiple candidates within a group of generated outputs rather than scoring each output independently.
Formally, within the GRPO framework, for a given input text $t_i$, the policy $\pi_\theta$ generates a group of $G$ candidate speech outputs $\{ s_{i,g} \}_{g=1}^{G}$. The model's parameters $\theta$ are then updated to maximize the following objective function:
\begin{equation}
\begin{split}
J_{\text{GRPO}}(\theta) =
& \ \mathbb{E}_{t_i \sim D,\ \{ s_{i,g} \}_{g=1}^{G} \sim \pi_{\theta_{\text{old}}}(\cdot \mid t_i)}
\left[ \frac{1}{G} \sum_{g=1}^{G} L_{i,g}(\theta) \right] \\
& \ - \beta D_{\text{KL}}(\pi_\theta \parallel \pi_{\theta_{\text{old}}}),
\end{split}
\label{eq:grpo_reward}
\end{equation}
where $L_{i,g}(\theta)=A_{i,g}^{G}\log \pi_\theta(s_{i,g}\mid t_i)$ is an advantage-weighted log-likelihood surrogate objective for the $g$-th candidate output of the $i$-th input. Here $A_{i,g}^{G}$ measures the performance of $s_{i,g}$ relative to the average quality of the group, so candidates with higher-than-average reward receive positive advantages and are assigned higher probability under $\pi_\theta$. The term $D_{\text{KL}}(\pi_\theta \parallel \pi_{\theta_{\text{old}}})$ is a KL divergence penalty that regularizes the policy update, preventing large deviations from the previous policy $\pi_{\theta_{\text{old}}}$ and ensuring training stability. The hyperparameter $\beta$ controls the strength of this regularization.

\subsection{Prosodic Label Alignment}
\label{sec:emotion_alignment}
The first component of the AI Feedback mechanism is prosodic-emotional label matching. To guide the model toward generating emotionally expressive speech, we adopt a fine-grained labeling strategy that annotates speech along four distinct prosodic-emotional dimensions: \textit{Structure, Emotion, Speed,} and \textit{Tone}. Here, \textit{Structure} captures the utterance-level discourse function,
\textit{Emotion} is a five-way category,
\textit{Speed} denotes speaking rate, and
\textit{Tone} describes pragmatic speaking style. These four labels represent key and complementary aspects of emotional expression, thereby forming a comprehensive yet manageable framework\cite{zhao2026stylebench}.

To implement this strategy at scale, we first generate target prosodic-emotional labels for the entire training dataset using an LLM\cite{ge2026sagelm}, and then optimize the policy with these labels as supervision. We define the match indicator as:
\begin{equation}
m_k(s_i)=\mathbb{I}\!\left[\hat{y}_{i,k}(s_i)=y_{i,k}\right], \quad k\in\mathcal{K},
\label{eq:mk_def}
\end{equation}
where $\mathcal{K}=\{\textit{Structure, Emotion, Speed, Tone}\}$ denotes the evaluation fields, each field $k\in\mathcal{K}$ has a discrete label set $\mathcal{Y}_k$, $y_{i,k}\in\mathcal{Y}_k$ is the target label for sample $i$, and $\hat{y}_{i,k}(s_i)\in\mathcal{Y}_k$ is the predicted label given speech $s_i$.

The reward function for this prosodic-emotional label alignment is formally defined as:
\begin{equation}
R_{\text{label}}(s_i) = \sum_{k\in\mathcal{K}} w_k \cdot m_k(s_i),
\label{eq:label_reward}
\end{equation}
where $m_k(s_i)$ is a binary match indicator for each field, and $w_k$ is the corresponding weight for that field.

\subsection{Semantic Accuracy Feedback}
\label{sec:text_alignment}
The second key component of the AI Feedback mechanism is designed to ensure speech accuracy. This is achieved by quantifying the semantic accuracy of the synthesized speech. Let $t_i$ be the original input text and $s_i$ be a generated speech sample. We employ an ASR model to obtain the transcription $\text{ASR}(s_i)$.

The accuracy-driven reward component, $R_{\text{wer}}$, is then formulated as the WER computed between the original text $t_i$ and the transcribed text $\text{ASR}(s_i)$:
\begin{equation}
R_{\text{wer}}(s_i) = \text{WER}\big(t_i, \text{ASR}(s_i)\big),
\label{eq:wer_reward}
\end{equation}
where $\text{WER}(\cdot)$ calculates the standard word error rate between the two input texts. This value serves as a direct cost within our composite reward function (Eq.~\ref{eq:total_reward}), effectively penalizing any deviation from the source text. By integrating this semantic accuracy penalty, the framework ensures that any improvements in emotional expressiveness do not come at the expense of clarity or content accuracy, leading to speech that is both articulate and emotionally resonant.

\section{Experimental Methodology}
In this section, we describe the datasets, baselines, evaluation metrics, and implementation details of our experiments.
\begin{table*}[t]
\centering
\caption{Objective and Subjective Evaluation Results Comparison of RLAIF-SPA with Baselines on WER, SIM-O, CMOS, Emotion MOS, and Speech Emotion Recognition across three datasets. \textbf{Bold} indicates the best result in each column, while \underline{underlining} indicates the second best.}
\label{tab:main_results}
\begin{tabular}{l|cc|cc|cccccc}
\noalign{\hrule height 0.3mm}
\multirow{2}{*}{\textbf{Model}} & \multicolumn{2}{c|}{\textbf{Objective}} & \multicolumn{2}{c|}{\textbf{Subjective}} & \multicolumn{6}{c}{\textbf{Speech Emotion Recognition}}\\

\multirow{-2}{*}{\textbf{Model}} & \textbf{WER$\downarrow$} & \textbf{SIM-O$\uparrow$} & \textbf{CMOS$\uparrow$} & \textbf{Emotion MOS$\uparrow$} & \textbf{Neutral$\uparrow$} & \textbf{Happy$\uparrow$} & \textbf{Sad$\uparrow$} & \textbf{Angry$\uparrow$} & \textbf{Surprise$\uparrow$} & \textbf{Avg$\uparrow$} \\

\hline
\rowcolor{gray!10}
\multicolumn{11}{c}{\textit{{LibriSpeech test-clean (En)}}} \\
\hdashline
Chat-TTS & 7.85 & 0.66 & 5.99 ± 0.56 & 5.75 ± 0.41 & 76.43 & 9.84 & \underline{37.31} & 1.47 & 0.00 & \underline{25.01}\\
F5-TTS & \textbf{4.87} & 0.70 & 6.97 ± 0.50 & 5.93 ± 0.36 & \underline{80.85} & 15.03 & 5.47 & \underline{7.35} & \underline{2.94} & 22.33\\
MegaTTS3 & 6.90 & \underline{0.71} & \underline{7.10 ± 0.39} & \underline{6.12 ± 0.44} & \textbf{82.00} & 8.29 & 9.45 & 5.88 & 0.00 & 21.12\\
Spark-TTS & 9.25 & 0.68 & 6.18 ± 0.58 & 5.73 ± 0.54 & 36.33 & \textbf{49.22} & 3.48 & \underline{7.35} & \underline{2.94} & 19.86\\
\textbf{RLAIF-SPA} & \underline{5.80} & \textbf{0.72} & \textbf{7.23 ± 0.40} & \textbf{6.53 ± 0.52} & \underline{80.85} & \underline{16.06} & \textbf{38.80} & \textbf{8.82} & \textbf{5.88} & \textbf{30.08}\\

\hline
\rowcolor{gray!10}
\multicolumn{11}{c}{\textit{{MELD (En)}}} \\
\hdashline
Chat-TTS & 19.64 & \underline{0.54} & 6.17 ± 0.46 & 5.85 ± 0.35 & 61.17 & 29.37 & 21.60 & 3.44 & 20.56 & 27.23\\
F5-TTS & \underline{9.07} & 0.44 & \underline{6.63 ± 0.43} & 6.10 ± 0.32 & \textbf{71.57} & 24.60 & 14.40 & \textbf{36.26} & 15.56 & \underline{32.48}\\
MegaTTS3 & 9.57 & 0.53 & 6.58 ± 0.39 & \underline{6.25 ± 0.33} & 67.13 & 22.22 & \underline{29.60} & 3.82 & 28.89 & 30.33\\
Spark-TTS & 15.54 & 0.53 & 6.18 ± 0.49 & 5.97 ± 0.41 & 27.79 & \textbf{50.79} & 9.60 & 2.29 & \textbf{38.33} & 25.76\\
\textbf{RLAIF-SPA} & \textbf{8.92} & \textbf{0.55} & \textbf{6.73 ± 0.45} & \textbf{6.30 ± 0.37} & \underline{68.65} & \underline{29.76} & \textbf{41.60} & \underline{5.73} & \underline{29.44} & \textbf{35.04}\\

\hline
\rowcolor{gray!10}
\multicolumn{11}{c}{\textit{{ESD (Zh)}}} \\
\hdashline
Chat-TTS & 6.70 & 0.67 & 6.06 ± 0.44 & 5.92 ± 0.34 & 63.14 & 6.35 & \underline{2.57} & 2.57 & \underline{2.86} & 15.50\\
F5-TTS & 4.01 & 0.69 & \underline{6.45 ± 0.44} & 6.06 ± 0.35 & \underline{87.71} & 0.00 & 0.00 & \textbf{10.29} & 0.57 & 19.71\\
MegaTTS3 & 3.86 & \underline{0.72} & 6.40 ± 0.41 & \underline{6.28 ± 0.40} & 54.00 & \textbf{51.14} & 0.00 & 0.86 & 0.00 & \underline{21.20}\\
Spark-TTS & \textbf{2.40} & 0.70 & 6.20 ± 0.41 & 5.99 ± 0.35 & 82.29 & 5.56 & 0.00 & 4.00 & 0.00 & 18.37\\
\textbf{RLAIF-SPA} & \underline{3.68} & \textbf{0.74} & \textbf{6.50 ± 0.48} & \textbf{6.29 ± 0.40} & \textbf{89.71} & \underline{7.14} & \textbf{4.57} & \underline{4.86} & \textbf{3.14} & \textbf{21.88}\\
\noalign{\hrule height 0.3mm}
\end{tabular}
\vspace{-0.4cm}
\end{table*}

\textbf{Datasets.} During training, we use a subset of 1{,}000 utterances from the LibriSpeech train-clean set \cite{librispeech}. We annotate each transcript with the same four structured attributes using GPT-4o following the labeling strategy in Sec.~\ref{sec:emotion_alignment} to ensure consistency between training supervision and evaluation. Since LibriSpeech has no emotion labels, all attribute labels are inferred from text by GPT-4o. Instead of uniform sampling, we select a more expressive subset by ranking utterances with an LLM-derived expressiveness score from the same labeler and keeping the top 1{,}000.

To ensure a comprehensive evaluation, we employ three datasets. For LibriSpeech test-clean, we annotate its transcripts with the same four attributes using GPT-4o to ensure consistency with the training supervision. MELD~\cite{poria-etal-2019-meld}, derived from the Friends TV series, provides a challenging conversational English setting characterized by complex and dynamic emotional interactions. ESD~\cite{zhou2022emotional} is used for Mandarin evaluation and contains parallel emotional speech from native speakers across five emotion categories.


\textbf{Baselines.} We compare RLAIF-SPA against four representative TTS baselines:
Chat-TTS, which is also used as the underlying speech synthesis engine in our system;
F5-TTS, a non-autoregressive flow-matching model based on Diffusion Transformers;
MegaTTS3, a latent diffusion transformer with sparse speech--text alignment;
and Spark-TTS, an LLM-based autoregressive TTS system built upon the Qwen2.5 backbone.
Our overall framework is built upon MiniCPM-o 2.6, with Chat-TTS instantiated as its speech synthesis component.

\textbf{Evaluation Metrics.} We assess model performance using both objective and subjective metrics to evaluate accuracy, speaker consistency, and emotional expressiveness. 

For objective evaluation, we consider three aspects. Intelligibility is measured using the \texttt{Word Error Rate (WER)}, obtained by transcribing synthesized speech with Whisper-Large-v3 and comparing it against the reference text. Speaker similarity is quantified using \texttt{SIM-O}, computed via WavLM-Large \cite{Chen_2022}, where scores range from $[-1, 1]$ and higher values indicate closer similarity between the generated speech and the prompt. Emotional correctness is evaluated using a Speech Emotion Recognizer, emotion2vec-large \cite{ma2023emotion2vecselfsupervisedpretrainingspeech}, which predicts an emotion label for each synthesized utterance. We report SER accuracy for each emotion category and the unweighted macro-average across the five categories.

For subjective evaluation, we conduct human listening tests to assess perceptual quality and emotional fidelity. Speech naturalness is evaluated using the \texttt{Mean Opinion Score (MOS)}, including \texttt{CMOS} for overall quality (covering clarity, naturalness, and high-frequency details) and \texttt{Emotion MOS} for perceived emotional similarity between synthesized and ground-truth speech. We report 95\% confidence intervals for both metrics using a two-way cluster bootstrap that jointly resamples raters and utterances, presenting results as mean $\pm$ CI half-width. In total, 50 randomly selected samples are evaluated, each by at least 20 listeners.

\textbf{Implementation Details.}
RLAIF-SPA is built upon the MiniCPM-O 2.6 model. During GRPO training, the reward signal is computed automatically: Whisper-Large-v3 is used to compute WER for accuracy assessment, while Qwen2-Audio evaluates alignment with the four prosodic-emotional labels. The corresponding weights are set to $\alpha_1=0.3$ for the WER penalty and $\alpha_2=0.7$ for the label-based reward, with uniform weights ($w_k$) applied across all label dimensions. All components are initialized from pre-trained MiniCPM-O 2.6 checkpoints. Training is conducted for 7 epochs using a learning rate of $5 \times 10^{-6}$.

\begin{table*}[t]
\centering
\caption{Ablation Study on Model Performance. Base denotes the original model without post-training, and w/o Label Reward removes the prosodic label reward during post-training while keeping GRPO.}

\label{tab:ablation_results}
\begin{tabular}{l|cc|cc|cccccc}
\noalign{\hrule height 0.3mm}
\multirow{2}{*}{\textbf{Model}} & \multicolumn{2}{c|}{\textbf{Objective}} & \multicolumn{2}{c|}{\textbf{Subjective}} & \multicolumn{6}{c}{\textbf{Speech Emotion Recognition}}\\

\multirow{-2}{*}{\textbf{Model}} & \textbf{WER$\downarrow$} & \textbf{SIM-O$\uparrow$} & \textbf{CMOS$\uparrow$} & \textbf{Emotion MOS$\uparrow$} & \textbf{Neutral$\uparrow$} & \textbf{Happy$\uparrow$} & \textbf{Sad$\uparrow$} & \textbf{Angry$\uparrow$} & \textbf{Surprise$\uparrow$} & \textbf{Avg$\uparrow$} \\

\hline
\rowcolor{gray!10}
\multicolumn{11}{c}{\textit{{LibriSpeech test-clean (En)}}} \\
\hdashline
\textbf{RLAIF-SPA} & \textbf{5.80} & \textbf{0.72} & \textbf{7.23 ± 0.40} & \textbf{6.53 ± 0.52} & \textbf{80.85} & 16.06 & \textbf{38.80} & \textbf{8.82} & \textbf{5.88} & \textbf{30.08}\\
\quad w/o Label Reward & 8.08 & 0.65 & 5.78 ± 0.38 & 5.31 ± 0.48 & 78.23 & \textbf{23.32} & 18.91 & 1.47 & 0.00 & 24.39\\
\quad Base (No Post-training) & 8.89 & 0.63 & 5.59 ± 0.28 & 5.33 ± 0.48 & 76.10 & 15.87 & 21.89 & 2.94 & 0.00 & 23.36\\

\hline
\rowcolor{gray!10}
\multicolumn{11}{c}{\textit{{MELD (En)}}} \\
\hdashline
\textbf{RLAIF-SPA} & \textbf{8.92} & \textbf{0.55} & \textbf{6.73 ± 0.45} & \textbf{6.30 ± 0.37} & 68.65 & \textbf{29.76} & \textbf{41.60} & \textbf{5.73} & \textbf{29.44} & \textbf{35.04}\\
\quad w/o Label Reward & 18.90 & 0.54 & 6.17 ± 0.41 & 5.52 ± 0.29 & \textbf{70.56} & 25.40 & 41.60 & 3.44 & 17.22 & 31.64\\
\quad Base (No Post-training) & 17.32 & 0.53 & 5.46 ± 0.26 & 5.90 ± 0.39 & 67.64 & 25.79 & 36.8 & 3.44 & 18.33 & 30.4\\

\hline
\rowcolor{gray!10}
\multicolumn{11}{c}{\textit{{ESD (Zh)}}} \\
\hdashline
\textbf{RLAIF-SPA} & \textbf{3.68} & \textbf{0.74} & \textbf{6.50 ± 0.48} & \textbf{6.29 ± 0.40} & 89.71 & \textbf{7.14} & 4.57 & \textbf{4.86} & \textbf{3.14} & \textbf{21.88}\\
\quad w/o Label Reward & 3.83 & 0.73 & 5.96 ± 0.29 & 5.66 ± 0.24 & 88.57 & 3.14 & 4.29 & 1.43 & 1.14 & 19.71\\
\quad Base (No Post-training) & 3.75 & 0.73 & 5.89 ± 0.35 & 6.05 ± 0.38 & \textbf{91.14} & 1.14 & \textbf{5.14} & 0.86 & 2.29 & 20.11\\

\noalign{\hrule height 0.3mm}
\end{tabular}
\vspace{-0.2cm}
\end{table*}

\begin{figure*}[t]
    \centering
    \captionsetup[subfigure]{justification=centering, singlelinecheck=false}
    \newcommand{\PlotScale}{0.9}
    \newcommand{\PlotH}{5.3cm}

    \begin{subfigure}[b]{0.36\linewidth}
        \centering
        \begin{minipage}[t][\PlotH][t]{\linewidth}
            \vspace{0pt}
            \centering
\begin{tikzpicture}[baseline=(current bounding box.north)]
\definecolor{cBase}{RGB}{220,240,241}
\definecolor{cFull}{RGB}{252,236,223}
\begin{axis}[
    width=\linewidth,
    height=5.75cm,
    xbar,
    bar width=7pt,
    enlarge y limits=0.1,
    axis lines*=box,
    xtick pos=bottom,
    ytick pos=left,
    major y tick style={draw=none},
    minor y tick num=0,
    symbolic y coords={pron,aux,prep,det,conj,adv,num},
    ytick=data,
    y dir=reverse,
    yticklabels={
        \shortstack{pronoun},
        \shortstack{aux/modal},
        \shortstack{preposition},
        \shortstack{determiner/poss},
        \shortstack{conjunction},
        \shortstack{adverb/neg},
        \shortstack{number}
    },
    yticklabel style={font=\fontsize{9}{11}\selectfont,inner sep=0pt,xshift=2.5pt},
    tick label style={font=\fontsize{9}{11}\selectfont},
    scaled x ticks=false,
    xticklabel style={
        /pgf/number format/.cd,
        fixed,
        fixed zerofill,
        precision=2
    },
    xmin=0.68, xmax=1.00,
    xtick={0.70,0.80,0.90,1.00},
    xmajorgrids=true,
    grid style={gray!15},
    legend style={
        at={(0.5,1.0)},
        anchor=south,
        yshift=-1.5pt,
        legend columns=2,
        font=\fontsize{9}{11}\selectfont,
        draw=none,
        fill=white,
        fill opacity=0,
        text opacity=1,
        /tikz/every odd column/.append style={column sep=2pt},
        /tikz/every even column/.append style={column sep=14pt},
    },
    legend image post style={xscale=0.55,yscale=1.2},
]
\addplot[
    fill=cBase, draw=black,
    bar shift=-3.5pt,
    nodes near coords,
    point meta=x,
    nodes near coords align={horizontal},
    every node near coord/.append style={font=\fontsize{6}{10}\selectfont,xshift=1pt,anchor=west},
    /pgf/number format/fixed,
    /pgf/number format/fixed zerofill,
    /pgf/number format/precision=2,
    forget plot
] coordinates {
    (0.854246,pron) (0.846000,aux) (0.856454,prep)
    (0.840102,det) (0.856863,conj) (0.802083,adv) (0.710843,num)
};
\addplot[
    fill=cFull, draw=black,
    bar shift=3.5pt,
    nodes near coords,
    point meta=x,
    nodes near coords align={horizontal},
    every node near coord/.append style={font=\fontsize{6}{10}\selectfont,xshift=1pt,anchor=west},
    /pgf/number format/fixed,
    /pgf/number format/fixed zerofill,
    /pgf/number format/precision=2,
    forget plot
] coordinates {
    (0.934094,pron) (0.942000,aux) (0.936068,prep)
    (0.916244,det) (0.898039,conj) (0.947917,adv) (0.867470,num)
};
\addlegendimage{area legend, fill=cBase, draw=none,fill opacity=1}
\addlegendentry{Base}
\addlegendimage{area legend, fill=cFull, draw=none,fill opacity=1}
\addlegendentry{RLAIF-SPA}
\end{axis}
\end{tikzpicture}
        \end{minipage}
        \captionsetup{margin={18mm,0mm}}
        \caption{Token-level accuracy attribution.}
        \label{fig:delta_acc_hon}
    \end{subfigure}\hspace{5mm}%
    \begin{subfigure}[b]{0.306\linewidth}
        \centering
        \begin{minipage}[t][\PlotH][t]{\linewidth}
            \vspace{0pt}
            \centering
\begin{tikzpicture}[baseline=(current bounding box.north)]
\definecolor{cBase}{RGB}{220,240,241}
\definecolor{cFull}{RGB}{252,236,223}

    \begin{axis}[%
        width=1.07\linewidth,
        height=5.75cm,
        ybar,%
        bar width=8pt,
        enlarge x limits=0.16,%
        axis lines*=box,%
        xtick pos=bottom,%
        major x tick style={draw=none},%
        minor x tick num=0,%
        ytick pos=both,%
        symbolic x coords={N,H,Sa,A,Su},%
        xtick=data,%
        xticklabels={%
            \shortstack{Neutral},%
            \shortstack{Happy},%
            \shortstack{Sad},%
            \shortstack{Angry},%
            \shortstack{Surprise}%
        },%
        xticklabel style={
            font=\fontsize{9}{11}\selectfont,
            rotate=25,
            anchor=east,
            inner sep=1pt
        },
        scaled y ticks=false,%
        yticklabel style={/pgf/number format/.cd,fixed,fixed zerofill,precision=2},%
        tick label style={font=\fontsize{9}{11}\selectfont},%
        ymin=0,%
        ymax=1.0,%
        ytick={0.00,0.25,0.50,0.75,1.00},%
        ymajorgrids=true,%
        grid style={gray!15},%
        legend style={at={(0.5,1.0)},
        anchor=south,
        yshift=-1.5pt,
        legend columns=2,font=\fontsize{9}{11}\selectfont,draw=none,fill=white,fill opacity=0,text opacity=1,
        /tikz/every odd column/.append style={column sep=2pt},
        /tikz/every even column/.append style={column sep=14pt},
        /pgf/number format/fixed,
        /pgf/number format/fixed zerofill,
        /pgf/number format/precision=2,},
        legend image post style={xscale=0.55,yscale=1.2}%
    ]%

    \addplot[%
        fill=cBase,draw=black,
        bar shift=-4pt,%
        nodes near coords,point meta=y,nodes near coords align={vertical},%
        every node near coord/.append style={font=\fontsize{6}{10}\selectfont,yshift=0pt,xshift=-3.3pt},%
        /pgf/number format/fixed,/pgf/number format/fixed zerofill,/pgf/number format/precision=2,%
        forget plot%
    ] coordinates {%
        (N,0.716915) (H,0.081000) (Sa,0.095476) (A,0.010309) (Su,0.048941)%
    };%

    \addplot[%
        fill=cFull,draw=black,
        bar shift=+4pt,%
        nodes near coords,point meta=y,nodes near coords align={vertical},%
        every node near coord/.append style={font=\fontsize{6}{10}\selectfont,yshift=0pt},%
        /pgf/number format/fixed,/pgf/number format/fixed zerofill,/pgf/number format/precision=2,%
        forget plot%
    ] coordinates {%
        (N,0.919499) (H,0.188342) (Sa,0.225514) (A,0.033549) (Su,0.112781)%
    };%

    \addlegendimage{area legend, fill=cBase, draw=none,fill opacity=1}
    \addlegendentry{Base}
    \addlegendimage{area legend, fill=cFull, draw=none,fill opacity=1}
    \addlegendentry{RLAIF-SPA}

    \end{axis}
\end{tikzpicture}
        \end{minipage}
        \caption{Per-emotion target-score gain.}
        \label{fig:delta_emotion}
    \end{subfigure}\hspace{0.00mm}%
    \begin{subfigure}[b]{0.306\linewidth}
        \centering
        \begin{minipage}[t][\PlotH][t]{\linewidth}
            \vspace{0pt}
            \centering
\begin{tikzpicture}[baseline=(current bounding box.north)]
\definecolor{cBase}{RGB}{220,240,241}
\definecolor{cFull}{RGB}{252,236,223}

\begin{axis}[
    width=1.07\linewidth,
    height=5.75cm,
    ybar,
    bar width=8pt,
    enlarge x limits=0.16,
    axis lines*=box,
    xtick pos=bottom,
    major x tick style={draw=none},
    minor x tick num=0,
    ytick pos=both,
    symbolic x coords={stru,emo,spe,tone},
    xtick=data,
    xticklabels={
        \shortstack{Structure},
        \shortstack{Emotion},
        \shortstack{Speed},
        \shortstack{Tone}
    },
    xticklabel style={
    font=\fontsize{9}{11}\selectfont,
    rotate=25,
    anchor=east,
    inner sep=1pt
},
    tick label style={font=\fontsize{9}{11}\selectfont},
    scaled y ticks=false,
    yticklabel style={
        /pgf/number format/.cd,
        fixed,
        fixed zerofill,
        precision=2
    },
    ymin=0, ymax=3.6,
    ytick={0.00,0.9,1.8,2.7,3.60},
    ymajorgrids=true,
    grid style={gray!15},
    legend style={
        at={(0.5,1.0)},
        anchor=south,
        yshift=-1.5pt,
        legend columns=2,
        font=\fontsize{9}{11}\selectfont,
        draw=none,
        fill=white,
        fill opacity=0,
        text opacity=1,
        /tikz/every odd column/.append style={column sep=2pt},
        /tikz/every even column/.append style={column sep=14pt},
        /pgf/number format/fixed,
        /pgf/number format/fixed zerofill,
        /pgf/number format/precision=2,
        },
    legend image post style={xscale=0.55,yscale=1.2},
]

\addplot[
    fill=cBase, draw=black, 
    bar shift=-4pt,
    nodes near coords,
    point meta=y,
    nodes near coords align={vertical},
    every node near coord/.append style={font=\fontsize{6}{10}\selectfont, yshift=0pt},
    /pgf/number format/fixed,
    /pgf/number format/fixed zerofill,
    /pgf/number format/precision=2,
    forget plot
] coordinates {
    (stru,2.7069) (emo,1.7205) (spe,3.2631)
    (tone,2.8951)
};

\addplot[
    fill=cFull, draw=black, 
    bar shift=+4pt,
    nodes near coords,
    point meta=y,
    nodes near coords align={vertical},
    every node near coord/.append style={font=\fontsize{6}{10}\selectfont, yshift=0pt,,xshift=3.3pt},
    /pgf/number format/fixed,
    /pgf/number format/fixed zerofill,
    /pgf/number format/precision=2,
    forget plot
] coordinates {
    (stru,0.8485) (emo,0.8388) (spe,1.0685)
    (tone,0.8547)
};

\addlegendimage{area legend, fill=cBase, draw=none,fill opacity=1,}
\addlegendentry{Base}
\addlegendimage{area legend, fill=cFull, draw=none,fill opacity=1,}
\addlegendentry{RLAIF-SPA}

\end{axis}
\end{tikzpicture}
        \end{minipage}
        \caption{Minimal-pair label intervention.}
        \label{fig:delta_label}
    \end{subfigure}

    \caption{Fine-grained analyses of the ablation study.
    (a) Token-level accuracy change based on ASR alignment.
    (b) Mean increase in target-emotion recognizer score, grouped by emotion.
    (c) Minimal-pair label interventions showing normalized changes on target vs.\ non-target dimensions.}
    \vspace{-0.6cm}
    \label{fig:ablation}
\end{figure*}

\section{Experimental Results}
\subsection{Overall Performance}
We evaluate the effectiveness of RLAIF-SPA by benchmarking it against four strong baselines: Chat-TTS, F5-TTS, MegaTTS3, and Spark-TTS. Table~\ref{tab:main_results} summarizes the comparative performance across three diverse datasets, detailing the results from general objective (WER, SIM-O) and subjective (CMOS, Emotion MOS) metrics, as well as Speech Emotion Recognition (SER) accuracy.

As shown in the results, RLAIF-SPA maintains high accuracy across testing scenarios, balancing semantic accuracy with expressive generation. While non-autoregressive baselines exhibit transcription accuracy on clean data, RLAIF-SPA demonstrates superior robustness in complex, emotionally dynamic contexts, where it achieves lower WER compared to competitive models. This stability is a direct consequence of our methodology, which explicitly incorporates a WER-based penalty into the AI Feedback mechanism. By optimizing for semantic alignment alongside emotional expressiveness, our model is guided to produce speech that preserves articulation precision even when modulating intense prosody.

Beyond accuracy, RLAIF-SPA excels in emotional expressiveness and overall speech quality. This advancement is principally driven by our fine-grained, label-driven reward component, which enables the model to precisely modulate prosodic nuances across dimensions such as Structure, Emotion, and Tone. The model's superiority is validated through a suite of evaluations. Objectively, it consistently achieves the highest speaker similarity across all datasets. Regarding Speech Emotion Recognition, while certain baselines may show competitive performance in specific emotion categories, RLAIF-SPA achieves the highest average accuracy across all testing sets. This indicates that our model possesses a superior generalized ability to synthesize a diverse range of emotions distinctly, rather than over-optimizing for a single emotional style. Subjectively, human listeners award RLAIF-SPA higher ratings for both overall naturalness and emotional fidelity. 

\subsection{Ablation Study}
We conduct an ablation to isolate the effect of structured feedback by comparing Base, w/o Label Reward, and RLAIF-SPA. The w/o Label Reward variant uses only the ASR-based fidelity reward during GRPO post-training, while RLAIF-SPA additionally includes attribute-wise label rewards. 
Table~\ref{tab:ablation_results} shows that optimizing with the ASR-based fidelity reward can improve transcription accuracy on clean read speech, but its effect is less consistent under domain shift. Adding attribute-wise label rewards yields the most reliable overall trade-off, improving expressiveness while keeping WER competitive.

To complement the aggregate metrics, we design three diagnostic analyses in Fig.~\ref{fig:ablation} that directly test the key claims of our method: the fidelity feedback improves accuracy, the attribute-wise reward strengthens the intended emotion, and the structured decomposition reduces cross-attribute entanglement. First, Fig.~\ref{fig:delta_acc_hon} breaks down transcription accuracy by token categories, showing broadly distributed gains rather than improvements concentrated in a single token type, which supports that the fidelity signal improves overall content preservation. Second, Fig.~\ref{fig:delta_emotion} reports increased recognizer confidence for the ground-truth emotion across classes, confirming that attribute-wise supervision effectively amplifies the intended affect. Finally, Fig.~\ref{fig:delta_label} evaluates minimal-pair label flips and observes lower non-target interference, demonstrating that decomposed feedback yields better disentanglement with reduced cross-attribute leakage.
\section{Conclusion and Further Work}
This paper presents RLAIF-SPA, a novel framework that autonomously optimizes for both emotional expressiveness and accuracy in speech synthesis. By employing an AI Feedback mechanism with GRPO to enforce fine-grained prosodic consistency and semantic accuracy, RLAIF-SPA significantly outperforms strong baseline models on the LibriSpeech,  MELD and ESD datasets. Crucially, our work demonstrates the feasibility of generating emotionally rich and highly intelligible speech without reliance on costly manual annotations, paving the way for more scalable and data-efficient emotional TTS systems. Future work will focus on refining the reward mechanism and assessing the framework's scalability across a broader range of acoustic environments and languages. Another promising direction involves modeling how a speaker's transient emotional state dynamically shapes prosody.

\section*{Acknowledgment}
This work was supported in part by the National Science Foundation of China (Nos. 62276056 and U24A20334), the Yunnan Fundamental Research Projects (No.202401BC070021), the Yunnan Science and Technology Major Project (No. 202502AD080014), the Fundamental Research Funds for the Central Universities (Nos. N25BSS054 and N25BSS094), and the Program of Introducing Talents of Discipline to Universities, Plan 111 (No.B16009).





\bibliographystyle{IEEEbib}
\bibliography{refs}

\end{document}